\documentclass[10pt,twocolumn,letterpaper]{article}

\usepackage[pagenumbers]{cvpr} 

\usepackage{graphicx}
\usepackage{amsmath}
\usepackage{amssymb}
\usepackage{booktabs}

\usepackage[pagebackref,breaklinks,colorlinks]{hyperref}

\usepackage[capitalize]{cleveref}
\crefname{section}{Sec.}{Secs.}
\Crefname{section}{Section}{Sections}
\Crefname{table}{Table}{Tables}
\crefname{table}{Tab.}{Tabs.}
\usepackage{xcolor}

\usepackage{amsmath,amsfonts,bm}

\def\eqref#1{equation~\ref{#1}}

\def\1{\bm{1}}

\DeclareMathAlphabet{\mathsfit}{\encodingdefault}{\sfdefault}{m}{sl}
\SetMathAlphabet{\mathsfit}{bold}{\encodingdefault}{\sfdefault}{bx}{n}

\definecolor{darkblue}{HTML}{000074}
\usepackage[page,header]{appendix}
\usepackage{titletoc}
\usepackage{graphicx}

\usepackage[utf8]{inputenc} 
\usepackage[T1]{fontenc}    
\usepackage{hyperref}       
\usepackage{url}            
\usepackage{booktabs}       
\usepackage{amsfonts}       
\usepackage{nicefrac}       
\usepackage{microtype}      
\usepackage{xcolor}         
\usepackage{algorithm}
\usepackage{algpseudocode}
\usepackage{graphicx}
\usepackage{rotating}
\usepackage{amsmath}
\usepackage{bbm}

\begin{document}

\title{CLR-GAM: Contrastive Point Cloud Learning with Guided Augmentation and Feature Mapping}

\author{Srikanth Malla\\
Samsung Semiconductor US\\
{\tt\small srikanth.m@samsung.com}
\and
Yi-Ting Chen\\
National Yang Ming Chiao Tung University\\
{\tt\small ychen@cs.nycu.edu.tw}
}
\maketitle

\begin{abstract}
Point cloud data plays an essential role in robotics and self-driving applications.
Yet, annotating point cloud data is time-consuming and nontrivial while they enable learning discriminative 3D representations that empower downstream tasks, such as classification and segmentation. 
Recently, contrastive learning-based frameworks have shown promising results for learning 3D representations in a self-supervised manner. 
However, existing contrastive learning methods cannot precisely encode and associate structural features and search the higher dimensional augmentation space efficiently. 
In this paper, we present CLR-GAM, a novel contrastive learning-based framework with Guided Augmentation (GA) for efficient dynamic exploration strategy and Guided Feature Mapping (GFM) for similar structural feature association between augmented point clouds.
We empirically demonstrate that the proposed approach achieves state-of-the-art performance on both simulated and real-world 3D point cloud datasets for three different downstream tasks, i.e., 3D point cloud classification, few-shot learning, and object part segmentation. 
%
\end{abstract}

\section{Introduction}
~\label{sec:intro}
3D understanding is of key importance in a wide range of applications including healthcare, medicine, entertainment, robotics, and human-machine interaction. 
%
%
Several 3D vision research problems (e.g., 3D point cloud classification~\cite{qi2017pointnet,qi2017pointnet++,wang2019dynamic}, detection~\cite{misra2021end}, and segmentation~\cite{qi2017pointnet++,thomas2019kpconv,wang2019dynamic}) have recently drawn much attention. 
However, obtaining 3D point cloud representations from the raw point clouds is challenging and often requires supervision, which causes high annotation costs.
\begin{figure}[t!]
    \includegraphics[width=0.9\columnwidth]{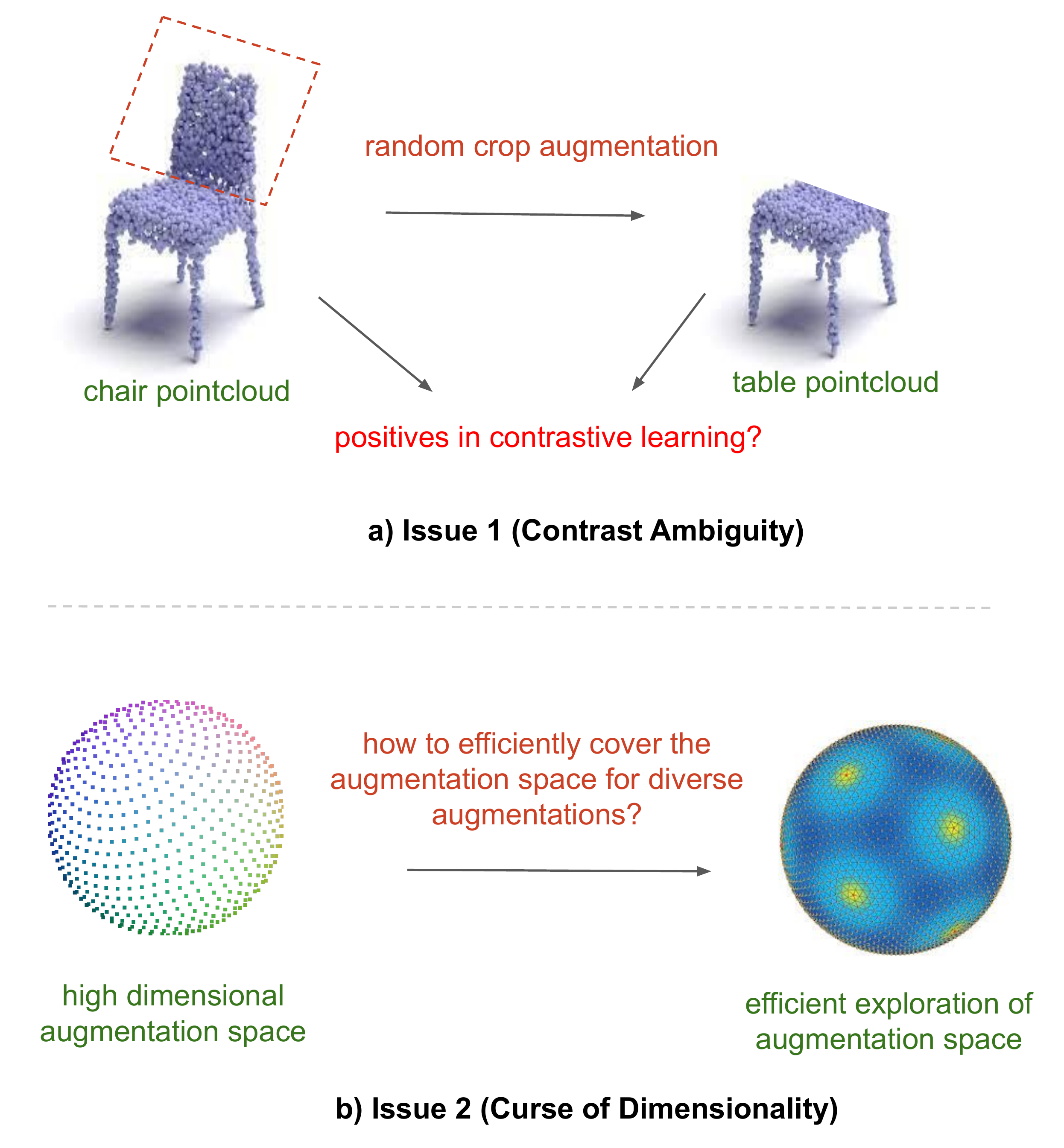}
    \caption{Motivation for CLR-GAM: a) motivation for Guided Feature Mapping, for better association b) motivation for Guided Augmentations, for better exploration of augmentation space}
    \label{fig:issues}
    \vspace{-0.3cm}
\end{figure}
As a result, self-supervised learning for 3D point cloud representations has witnessed much
progress and can potentially improve sample efficiency and generalization for these 3D understanding tasks.
Existing works are mainly based on generative models~\cite{achlioptas2018learning, han2019view,wu2016learning}, reconstruction~\cite{eckart2021self,han2019multi, li2018so, yang2018foldingnet, zhao20193d}, pretext task~\cite{wang2021unsupervised, poursaeed2020self, sauder2019self, hassani2019unsupervised, sun2021point, yang2021progressive, rao2020global}, and contrastive learning~\cite{ zhang2019unsupervised, sanghi2020info3d, xie2020pointcontrast,huang2021spatio, liu2021point, zhang2021self,du2021self}.
%
Much progress has been made in recent contrastive learning-based methods. 
%
%
However, we observe the following two limitations. 

\noindent\textbf{Issue 1 (Contrast Ambiguity):} 
\textbf{a) GCA (Global Contrast Ambiguity).} With augmentations like cropping and nonrigid body transformation, the shape of an augmented object is entirely different from the original object, leading to ambiguity for contrastive learning. For instance, if we remove the back part of a "Chair" point cloud, the resulting point cloud could be similar in shape to a sample of the "Table" class, as shown in Figure~\ref{fig:issues}.a. It poses a challenge for contrastive learning based methods because they do not access class labels for training. \textbf{b) LCA (Local Contrast Ambiguity).} In addition, local feature contrasting techniques~\cite{xie2020pointcontrast,liu2021point} treat every other point's feature in the same point cloud as a negative. The drawback with this objective is that there are symmetries and similar shapes in an object that can have the same features.
%

\noindent\textbf{Issue 2 (Curse of Dimensionality):} contrastive learning requires a variety of augmentations to learn discriminative 3D point cloud representations.
However, searching over these high-dimensional augmentations is time-consuming and does not guarantee proper coverage with a dynamic limited number of samples. 

In this work, we introduce two novel modules, i.e., guided feature mapping (GFM) and guided augmentation (GA), to overcome the above limitations.
We introduce the GFM module to associate features of the same structure between two augmented samples for effective feature association under heavy shape deformation. The feature contrasting is done at the object or global level, like most works, but with a tight coupling of local feature association.  
The GA module is present to efficiently explore higher-dimensional augmentation spaces with dynamically limited samples for diverse coverage of the augmentation space. 
We conducted extensive experiments to validate the effectiveness of the proposed contrastive learning framework. 
Specifically, we benchmark three downstream tasks, i.e., classification, few-shot learning, and object part semantic segmentation.
We obtain state-of-the-art performance on the three tasks, and extensive ablative studies are conducted to justify the designed choice. 

\noindent\textbf{Our main contributions:} i) We propose Guided Augmentation (GA) and Feature Mapping (GFM) for learning discriminative 3D point cloud representations. ii) Our proposed approach achieves state-of-the-art performance on three downstream tasks, i.e., object classification, few-shot learning, and part segmentation. iii) Extensive ablatives studies are presented to justify our design choices.



\section{Related Works}

\subsection{Contrastive Learning on Point Clouds}
\vspace{-0.3cm}
\begin{table}[h!]
\centering
\resizebox{0.95\columnwidth}{!}{
\begin{tabular}{ c|c|c|c|c } 
Method&\multicolumn{2}{c|}{Feature Contrast}&\multicolumn{2}{c}{Contrast Ambiguity}\\\hline
Contrastive&global contrast & local contrast& GCA & LCA \\
\hline
PointContrast~\cite{xie2020pointcontrast}&&\checkmark& &\checkmark\\ 
PointDisc~\cite{liu2021point}&&\checkmark &&\checkmark\\
DepthContrast~\cite{zhang2021self}&\checkmark& &\checkmark&\\ 
STRL~\cite{huang2021spatio}&\checkmark& &\checkmark&\\ 
CrossPoint~\cite{afham2022crosspoint}&\checkmark& &\checkmark&\\\hline 
\textbf{CLR-GAM (ours)}&\checkmark& & & \\ 
\end{tabular}
}
\caption{Comparison of existing works and the problems}
\label{tbl:comp_related_works}
\vspace{-0.2cm}
\end{table}
Following the recent success of self-supervised contrastive learning for images, recent works~\cite{du2021self, huang2021spatio, liu2021point, sanghi2020info3d, xie2020pointcontrast, zhang2019unsupervised, zhang2021self} explore contrastive learning for point cloud. 
PointContrast~\cite{xie2020pointcontrast} applies contrastive loss for pointwise features generated from the neural network for a point cloud transformed using two random augmentations, to learn invariant features. PointContrast uses local feature contrasting, whereas in our approach, we tightly couple local feature association with object-level/global feature contrasting. Most importantly, the features of different points in the same object can be similar because of symmetries and similar shapes, but PointContrast treats every other point's feature in the same pointcloud as a negative feature and suffers from LCA as shown in Table~\ref{tbl:comp_related_works}. 
DepthContrast~\cite{zhang2021self} uses two encoders for global level contrasting using voxel and point encoders but does not address GCA. 
Zhu et al.~\cite{zhu2021improving} uses the feature memory bank~\cite{he2020momentum} to store negatives and positives for hard sample mining. 
Huang et al.~\cite{huang2021spatio} propose STRL that applies spatial augmentation for temporally correlated frames in a sequence point cloud dataset and performs contrastive learning. 
%
%
Recently, Afham et al.~\cite{afham2022crosspoint} proposed CrossPoint that learns cross-modal representations (images and point clouds) using contrastive learning.
All these methods rely on contrastive learning of the encoded global features of point clouds, ignoring the structural deformations that lead to intraclass confusion (GCA). 
Recently, the authors of PointDisc~\cite{liu2021point} apply a point discrimination loss within an object to enforce similarity in features for points within a local vicinity. 
PointDisc makes the geometric assumption of a fixed radius for obtaining positives from the encoded features of the same point cloud and also suffers from LCA, similar to PointContrast.
In this work, we introduce the GFM to identify structurally similar features between two different augmentations of the same point cloud without any geometric assumptions.
We empirically demonstrate the effectiveness of the proposed GFM to learn discriminative 3D representations for three different downstream tasks.
\vspace{-0.2cm}
\subsection{Guided Augmentation}
Several guided augmentation approaches for image modality~\cite{charalambous2016data, hauberg2016dreaming, rogez2016mocap, peng2015learning, dixit2017aga} have shown to synthesize variable realistic samples for training. 
It is an important problem to generalize an algorithm to cover the unseen samples in the test data, which is expected to have wide variations of augmentation. 
In the context of human posture, \cite{charalambous2016data} generates synthetic videos for gait recognition, and \cite{rogez2016mocap} augments images with 2D poses using 3D MoCAP data for pose estimation. 
For improving image detection,~\cite{peng2015learning, su2015render} renders 3D CAD models with variable texture, background, and pose for generating synthetic images.
Hauberg et al.\cite{hauberg2016dreaming} learn class-specific transformations (diffeomorphism) from external data, whereas another work ~\cite{miller2000learning} synthesizes new images using an iterative process.
Since the existing works are task-specific and designed for supervised learning of image modality, they require class labels during training. AGA~\cite{dixit2017aga} extends to the feature space to be class agnostic, but it requires a huge corpus of annotated datasets with class labels to pretrain. We cannot directly adapt those approaches to self-supervised point cloud learning approaches, so we find exploration strategies in reinforcement learning are relevant for unsupervised guided augmentation.

\subsection{Exploration of High Dimensional Spaces}
Efficient exploration in high-dimensional space is a fundamental problem in reinforcement learning. 
Different strategies such as selecting new states including epsilon-greedy, selecting random states with epsilon probability~\cite{mnih2015human}, upper confidence bounds~\cite{auer2002using}, Boltzmann exploration~\cite{watkins1989learning, sutton1990integrated} using softmax over the utility of actions and Thomson sampling~\cite{agrawal2012analysis}. 
The motivation or curiosity to explore new states is coined as intrinsic motivation~\cite{oudeyer2008can}, which is adapted into ~\cite{bellemare2016unifying, haber2018learning, houthooft2016vime, oh2015action, ostrovski2017count, pathak2017curiosity, stadie2015incentivizing} as an intrinsic reward to quantify how different the new state is from already explored states. 
Some existing methods~\cite{haber2018learning, houthooft2016vime, oh2015action, pathak2017curiosity, stadie2015incentivizing} use error in prediction as an intrinsic reward, while others use count-based techniques~\cite{ostrovski2017count, bellemare2016unifying}. 
However, the computation of intrinsic reward using function approximation is slow to catch up and is not efficient enough for contrastive learning. 
%
In this work, we introduce a guided augmentation mechanism for efficient exploration of new states using a memory-based module motivated by~\cite{badia2020never}. Badia et al. construct an episodic memory-based intrinsic reward using k-nearest neighbors over the explored states to train the directed exploratory policies. 


\section{Methodology}
\begin{figure*}[t]
    \centering
    \includegraphics[width=0.8\textwidth]{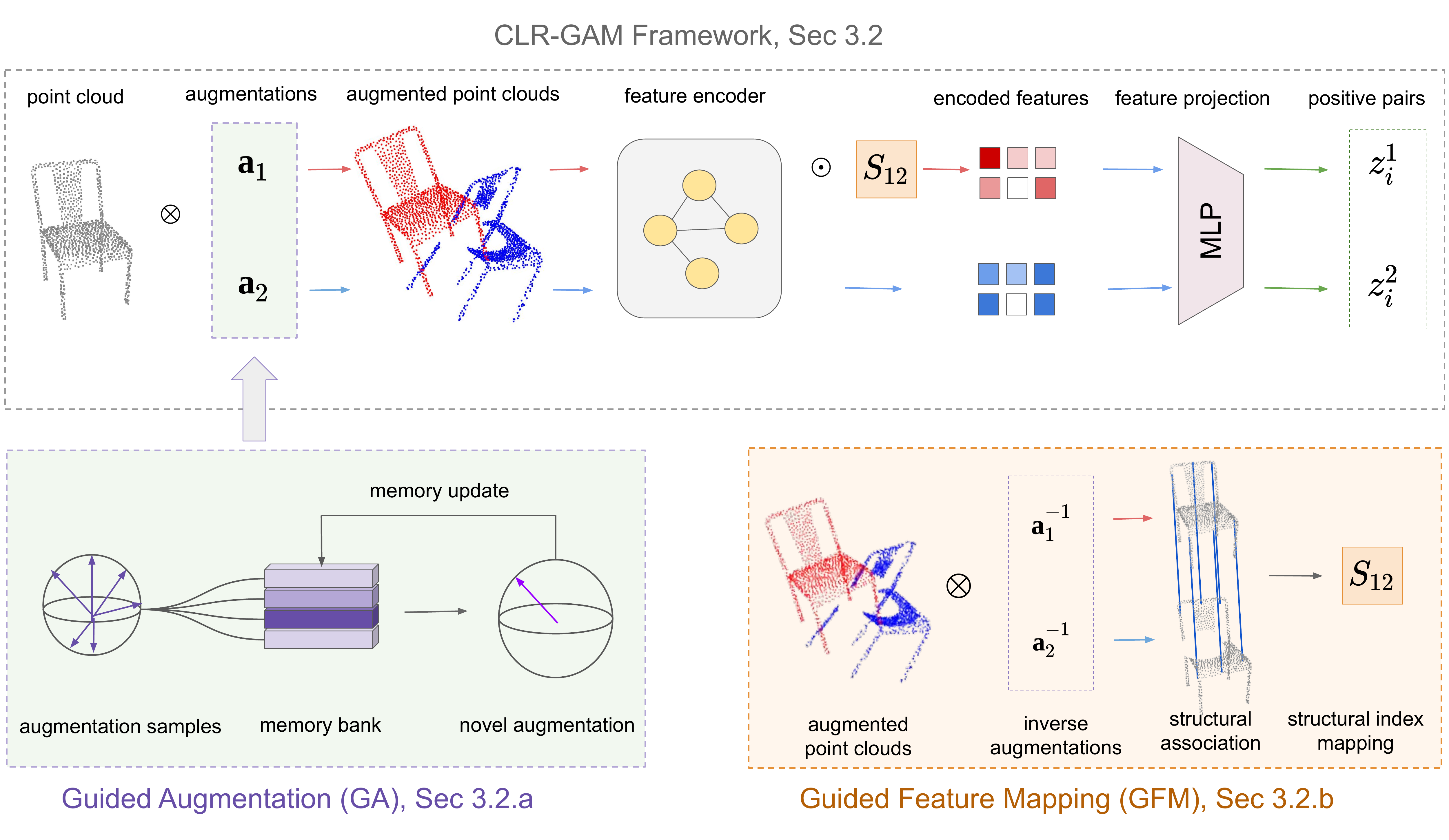}
    \vspace{-0.3cm}
    \caption{The proposed CLR-GAM framework with guided augmentation (GA) and guided feature mapping (GFM). $\otimes$ is the augmentation operator, $\odot$ is the indexing operator and $S_{12}$ is the structural index mapping.}
    \label{fig:motivation}
    \vspace{-0.6cm}
\end{figure*}
\subsection{Preliminaries and Notation}
We denote a point cloud as $P_i$, which consists of an unordered set of points $\mathbf{x}_{j=1:n}$ and $\mathbf{x}_j\in \mathbb{R}^3$, where the parameter $n$ is the number of points, and a point $\mathbf{x}_j$ is in 3D coordinate space. 
A point cloud $P_i$ can be augmented by changing scale $\mathbf{a}^S_k\in \mathbb{R}^3$, translation $\mathbf{a}^T_k\in \mathbb{R}^3$, rotation $\mathbf{a}^R_k\in \mathbb{R}^3$, cropping $\mathbf{a}^C_k$, and jittering $\mathbf{a}^J_k$. 
The combined set of the above operations is denoted as $\mathbf{a}_{k}$, where $\mathbf{a}_{k} = [\mathbf{a}^C_k, \mathbf{a}^S_k, \mathbf{a}^R_k, \mathbf{a}^T_k, \mathbf{a}^J_k]$.
Given a point cloud $P_i$, we apply the order defined in $\mathbf{a}_{k}$ to obtain an augmented point cloud $P^k_i$. 
In the remaining of this paper, we use $i,j,k$ as the index of a point cloud $P_i \in \mathbb{R}^{n\times3}$ and the corresponding encoded features $F_i \in \mathbb{R}^{n\times d}$, a point in point cloud $x_j = P_i(j)\in \mathbb{R}^{1\times3}$ and a row of the encoded features $F_i(j) \in \mathbb{R}^{1\times d}$, and an augmentation operation $\mathbf{a}_{k}$, respectively. Note that the parameter $n$ is the number of points in a point cloud.
\vspace{-0.2cm}
\subsection{Framework}
The detailed architecture of the CLR-GAM framework, a contrastive learning-based approach with the proposed GA and GFM modules, is depicted in Figure~\ref{fig:motivation}. 
We briefly introduce the overall contrastive learning algorithm in this section.  
First, a point cloud $P_i$ is transformed into $P^1_i$ and $P^2_i$ by applying two augmentation operations $\mathbf{a}_1$ and $\mathbf{a}_2$.
%
%
We utilize a Siamese architecture with shared weights for feature extraction.
In this work, we utilize PointNet (a MLP based method)~\cite{qi2017pointnet} and DGCNN (a graph convolution-based method)~\cite{wang2019dynamic} to extract features that are invariant to the input order. 
The augmented point clouds $P^1_i, P^2_i \in \mathbb{R}^{n\times3}$ are encoded into latent space $F_i^1, F_i^2 \in \mathbb{R}^{n\times d}$, respectively.
The parameter $n$ is the number of points, and $d$ is the feature dimension.
%
%
%
%
%
The augmented point clouds $P^1_i$, $P^2_i$ could contain different structures, while both point clouds originate from the same point cloud $P_i$.
To ensure an effective feature association between $F^1_i$ and $F^2_i$, we introduce the Guided Feature Mapping (GFM) module to associate the features that belong to the same structure between two augmented point clouds. 
The feature $F^{1}_i$ is mapped to $F^{12}_i$ to entail similar structural features when $F^{2}_i$ is considered. 
%
The features $F^{12}_i$ and $F^2_i$ are pooled and projected into the projected latent space, resulting in $z^1_i$ and $z^2_i$, respectively. 
We perform contrastive loss to enforce that the latent representation distance between the same point clouds (positives) features is smaller than the distance between the features from different point clouds (negatives) in a minibatch.
%
%
In addition, contrastive learning heavily relies on the quality of augmentation. 
An efficient strategy for exploring the augmentation space is indispensable.
%
%
We introduce a guided augmentation search to explore various augmentations efficiently, motivated by~\cite{badia2020never}.
%
%

\textbf{a) Guided Augmentation:}
\label{sec:ga}
Augmentation is the key to the success of self-supervised contrastive learning.
We hypothesize that if we can efficiently identify a wide range of informative augmentations, a discriminative representation can be learned.
Existing approaches apply random sampling in augmentation spaces, which leads to ineffective augmentation and a high computational burden.
Thus, we utilize a dynamic and efficient exploration strategy commonly used in reinforcement learning to mitigate the limitation.

The ranges of each dimension of rotation $\mathbf{a}^R$, translation $\mathbf{a}^T$, and scaling $\mathbf{a}^S$ are $[0, 2\pi)$ radians, $[-1 , 1]$ meters, and $[0.5 , 1]$, respectively. 
%
%
Since the jittering and cropping operations are point specific, we ignore them in guided augmentation for simplicity. 
%
%
%
 Specifically, motivated by~\cite{badia2020never}, we utilize a memory bank $M$ to save explored augmentation samples $\mathbf{a}_m$, where $m$ is the index of a slot. 
 The goal is to ensure that the new sample is different from the explored samples.
 It is worth noting that it is hard to obtain this behavior when just the average of $L$-norm distance is used to select novel augmentations.  
 %
 %
 To start, we first randomly sample $N$ augmentations $\hat{a}_{k=1:N}$ from the augmentation space $\mathbf{a}_k$.
 %
%
 We compute the distance of a new sample $\hat{\mathbf{a}}_k$ from all the explored samples in the memory bank $\mathbf{a}_m$.
 The design is used to evaluate the novelty of a sample.
 %
%
  A novel augmentation $\mathbf{a}^*_k$ is identified by using equation~\ref{eq:inverse_dirac}.
  %
  %
\begin{equation}
\mathbf{a}^*_k=\arg_{\hat{a}_k}\max \frac{1}{\sqrt{\sum_{m \in M} K(\mathbf{a}_m , \hat{\mathbf{a}}_k)}+c}
\label{eq:inverse_dirac}
\end{equation}
where $K(\mathbf{a}_m, \mathbf{a}_k) =\frac{\epsilon}{d(\mathbf{a}_m,\mathbf{a}_k)+\epsilon}$.
The distance function $d$ between two augmentations is the $L_2$-norm.
 The parameters $c,\epsilon$ are small values added for numerical stability. 
 The memory bank is updated with the selected novel augmentation $\mathbf{a}^*_k$. 
 The operation is applied twice on each point cloud $P_i$ in an iteration to obtain two novel augmentations $\mathbf{a}_1, \mathbf{a}_2$. The two augmentations are applied to input point cloud $P_i$, as shown in Figure~\ref{fig:motivation}. 
%
Note that the augmentations of rotation angles $2\pi$ and $0$ are the same in the angular space, we utilize an angular distance measure, i.e., $d_R(\mathbf{a}^R_m,\mathbf{a}^R_n)= \sum ( 0.5 - |\ |\textbf{a}^R_m - \textbf{a}^R_n| - 0.5|)$, instead of using $L_2$ distance.
%
To be consistent with different scales and ranges of augmentations, we normalize each augmentation to $[0,1]$ before computing the total distance $d$ as shown in equation~\ref{eq:total_distance}, where $\alpha_{R}$, $\alpha_{T}$, and $\alpha_{S}$ are the weights for the three distances.
%
\begin{equation}
\begin{split}
d(\mathbf{a}_m,\mathbf{a}_n)=&\alpha_R d_R(\mathbf{a}^R_m,\mathbf{a}^R_n) +\alpha_T||\mathbf{a}^T_m-\mathbf{a}^T_n||_2\\&+\alpha_S||\mathbf{a}^S_m-\mathbf{a}^S_n||_2
\label{eq:total_distance}
\end{split}
\end{equation}


\textbf{b) Guided Feature Mapping:}
\label{sec:gfm}
To learn discriminative point cloud representations, it is crucial to project features with similar structural characteristics for contrastive learning. 
%
%
Existing methods may fail to identify the structural similarity between the two augmented point clouds because certain augmentations (e.g., cropping, scaling) could lead to heavy deformations of an augmented point cloud with a completely different shape from the original class and similar to a different class.
Based on our observation, when both the augmentations $\mathbf{a}_1, \mathbf{a}_2$ contains crop operations, this results in the very limited structural similarity between the augmented point clouds. So we exclude the crop augmentation $\mathbf{a}^C_1$ from the augmentation $\mathbf{a}_1$. 
%
In $\mathbf{a}_2$, it uses all the augmentations, i.e., rotation, translation, scaling, cropping, and jittering.
%
Note that $\mathbf{a}^R_k,\mathbf{a}^T_k,\mathbf{a}^S_k$ are invertible operations as they are applied on the whole point cloud. The operation $\mathbf{a}^J_k$ is a point-specific operation and invertible. 
On the other hand, the cropping operation $\mathbf{a}^C_k$ is not invertible as the information is lost. 
An invertible augmentation operation can be written as $P_i= (\mathbf{a}_1)^{-1}\otimes P^1_i$, where $P^1_i$ is an augmented point cloud, $P_i$ is the original point cloud, and $\otimes$ denotes an augmentation operator.
%
%
%
The equation holds because the augmentation $\mathbf{a}_1$ does not contain a cropping operation.
Whereas the augmentation inverted point cloud of $P^2_i$ results in $P^C_i= (\mathbf{a}_2)^{-1}\otimes P^2_i$, a cropped point cloud. The crop operation is ignored in the inverse operation with $\mathbf{a}_2$, as it is not invertible.
The order of points and their structures cannot be directly associated with these two augmented point clouds even with the same number of points. 
The closest point association mapping $S_{12}$ between points of inverted point clouds of $P^1_i$ and $P^2_i$ is calculated based on equation~\ref{eq:association}. 
The structural index mapping $S_{12}$ retains only the indices of the closest points of $P^1_i$ to $P^2_i$, for every point in $P^2_i$ with index $j$.
\begin{equation}
    S(j)_{12}= \arg_q \min || P^C_i(j)-P_i(q)||_2
\label{eq:association}
\end{equation}

%
The operators $P_i(\cdot)$ and $F_i(\cdot)$ denote an indexing operation to point cloud and feature set, respectively.
The guided mapped feature $F^{12}_i$ is obtained according to $F^{12}_i= F^{1}_i(S_{12})$. 
The feature $F^{12}_i$ is projected to $z^1_i$ using the feature projection module after pooling. Feature projection module is an MLP to reduce the dimensionality of the features.
Similarly, $F^2_i$ is projected to $z^2_i$.
The contrastive loss~\cite{chen2020simple} is utilized to compute the similarity between positives ($z^1_i, z^2_i$) and negatives from the minibatch. We do not store negatives over multiple iterations in a memory bank for comparability with other techniques~\cite{afham2022crosspoint}, which is commonly done for improving the performance~\cite{he2020momentum}.
The loss can be found in equation~\ref{eq:contrast_loss}.
%
%
The similarity measure is the cosine distance between two features, $\textrm{sim}(z_1,z_2)=(z_1^Tz_2)/(||z_1|| ||z_2||)$. 
Given a minibatch, the final contrastive loss is $L_c=\frac{1}{2B}\sum_{b=1}^B(L^b_{\mathbf{1,2}}+ L^b_\mathbf{2,1})$.
%
The parameter $\tau$ is temperature 0.5, $b$ is the index of the feature in the minibatch of total size $B$. 
%
\begin{equation}
\resizebox{\columnwidth}{!}{
    $L^i_\mathbf{1,2} = -log\frac{\textrm{exp}(\textrm{sim}(z^{1}_i, z^{2}_i)/\tau)}{\sum^B_{b=1, b\neq i} \textrm{exp}(\textrm{sim}(z^{1}_i, z^{1}_b)/\tau)+\sum^B_{b=1} \textrm{exp}(\textrm{sim}(z^{1}_i, z^{2}_b)/\tau)}$
}
\label{eq:contrast_loss}
\end{equation}
      

\section{Experiments}
In this section, we first quantitatively evaluate the self-supervised trained approach on different downstream tasks and different object data sets (synthetic and real world). Second, we qualitatively visualize the features on an unseen object dataset. Finally, we do ablation studies of a) our novel modules and augmentations, b) t-SNE feature visualization on the unseen dataset, and c) qualitatively visualize the features on an unseen driving scenario.
\subsection{Quantitative Results}
\begin{table}[h!]
\centering
\resizebox{0.87\columnwidth}{!}{
\begin{tabular}{l l c c}
Modality&Method &\multicolumn{2}{c}{ModelNet-40}\\\hline\hline
point cloud &3D-GAN~\cite{wu2016learning} & \multicolumn{2}{c}{83.3}\\
&Latent-GAN~\cite{achlioptas2018learning}& \multicolumn{2}{c}{85.7}\\
&SO-Net~\cite{li2018so} & \multicolumn{2}{c}{87.3}\\
&FoldingNet~\cite{yang2018foldingnet}& \multicolumn{2}{c}{88.4}\\
&MRTNet~\cite{gadelha2018multiresolution} & \multicolumn{2}{c}{86.4}\\
&3D-PCapsNet~\cite{zhao20193d}& \multicolumn{2}{c}{88.9}\\
&ClusterNet~\cite{zhang2019unsupervised} &\multicolumn{2}{c}{86.8}\\
&VIP-GAN~\cite{han2019view}& \multicolumn{2}{c}{90.2}\\
+ Image Modality&DepthContrast~\cite{zhang2021self}& \multicolumn{2}{c}{85.4}\\\hline
&&PNet& DGCNN\\\hline
point cloud&Multi-Task~\cite{hassani2019unsupervised}&-&89.1\\
&PoinDisc~\cite{liu2021point}&86.2&89.3\\
&Self-contrast~\cite{du2021self}&-&89.6\\
&PoinContrast~\cite{xie2020pointcontrast}&86.7&89.9\\
&Jigsaw~\cite{sauder2019self}&87.3&90.6\\
&STRL~\cite{huang2021spatio}&88.3&90.9\\
&Rotation~\cite{poursaeed2020self} &88.6&90.8\\
&OcCo~\cite{wang2021unsupervised}&88.7&89.2\\
&\textbf{CLR-GAM (ours)}&\textbf{88.9}&\textbf{91.3}\\\hline
+ Image Modality&CrossPoint~\cite{afham2022crosspoint}&\underline{89.1}&91.2\\

\end{tabular}
}
\caption{We pretrained using the proposed contrastive self-supervised learning framework on ShapeNet. We evaluate on the test split of ModelNet-40 by fitting a linear SVM classifier. The reported results are the overall accuracy. The upper subtable uses custom backbone and training strategies.}
\label{tbl:classification_mn40_results}
\vspace{-0.4cm}
\end{table}
\begin{table}[h!]
\centering
\resizebox{0.6\columnwidth}{!}{
\begin{tabular}{ c|c|c } 
Method & PNet & DGCNN \\\hline
Jigsaw~\cite{sauder2019self} & 55.2 & 59.5\\
PoinDisc~\cite{liu2021point} & 68.3 & 78.0\\
OcCo~\cite{wang2021unsupervised} & 69.5 & 78.3\\
PointContrast~\cite{xie2020pointcontrast} & 70.4 & 78.6\\
STRL~\cite{huang2021spatio} &74.2 & 77.9\\
\textbf{CLR-GAM (ours)}&\textbf{75.7}&\textbf{82.1}\\ \hline
CrossPoint~\cite{afham2022crosspoint}&75.6&81.7\\
\end{tabular}
}
\caption{3D Object classification on ScanObjectNN. We pretrained using the proposed contrastive self-supervised learning framework on ShapeNet. We evaluate on test split of ScanObjectNN by fitting a linear SVM classifier. The reported results are the overall accuracy on the test split.}
\label{tbl:classification_scanobject_resutls}
\vspace{-0.5cm}
\end{table}
\noindent\textbf{a) 3D Object Classification:} For this task, we utilize the ModelNet-40 (synthetic) and ScanObjectNN (real-world) datasets. 
The ModelNet-40 dataset consists of a wide range of 3D objects' CAD models.
%
The dataset contains 12,331 objects that are categorized into 40 classes. 
We use 9,843 for training and 2,468 for testing. 
The ScanObjectNN dataset is challenging because data is collected in cluttered environments, so objects could be partially observable due to occlusions.  
It consists of 15 classes totaling 2,880 objects (2,304 for training and 576 for testing).

We follow the same evaluation strategy as in the existing works~\cite{huang2021spatio, afham2022crosspoint, wang2021unsupervised}.
Specifically, we freeze the pretrained point cloud feature extractor pretrained on the ShapeNet dataset. %
We randomly sample 1024 points from each object for testing classification accuracy on ModelNet-40 and ScanObjectNN.
%
We fit a linear SVM~\cite{cortes1995support} on the extracted features.
The results on the testing set of ModelNet-40 and ScanObjectNN can be found in Table~\ref{tbl:classification_mn40_results} and Table~\ref{tbl:classification_scanobject_resutls}, respectively. 
Additionally, we also conduct experiments using two different backbones, i.e., PNet~\cite{qi2017pointnet} and DGCNN~\cite{wang2019dynamic}, on the two datasets. 
We demonstrate state-of-the-art performance on the ModelNet-40 dataset using both backbone architectures compared to point cloud pretrained approaches in the bottom sub-table, as shown in Table~\ref{tbl:classification_mn40_results}. 
With the DGCNN backbone, the proposed approach performs better than CrossPoint and DepthContrast. 
It is worth noting that both methods utilize extra image modality for pretraining, while the proposed contrastive self-supervised learning framework only uses point clouds. Compared to the previous SOTA on a single modality (OcCo), the accuracy is improved by 2.35\% (with DGCNN).

The results conducted on ScanObjectNN further justify the effectiveness of the proposed framework, as shown in Table~\ref{tbl:classification_scanobject_resutls}.
State-of-the-art performance is present compared to both point cloud and multimodal pretrained approaches using both backbone architectures. Noticeably, compared to previous SOTA on a single modality (OcCo), the accuracy is improved by 4.8\% (with DGCNN).
In addition to satisfactory results, we empirically demonstrate that the proposed approach has better generalization capability in a real-world setting under severe occlusions than other methods.
%
\begin{table*}[t!]
\centering
\resizebox{0.8\textwidth}{!}{
\begin{tabular}{ l | c c c c | c c c c}
&\multicolumn{4}{c|}{5-way}&\multicolumn{4}{c}{10-way}\\\hline
Method &\multicolumn{2}{c}{10-shot} &\multicolumn{2}{c|}{20-shot} &\multicolumn{2}{c}{ 10-shot} &\multicolumn{2}{c}{20-shot}\\\hline\hline
FoldingNet~\cite{yang2018foldingnet}&\multicolumn{2}{c}{33.4±4.1} & \multicolumn{2}{c|}{35.8±5.8} &\multicolumn{2}{c}{18.6±1.8}&   \multicolumn{2}{c}{15.4±2.2}\\
Latent GAN~\cite{achlioptas2018learning}&\multicolumn{2}{c}{41.6±5.3} & \multicolumn{2}{c|}{46.2±6.2} &\multicolumn{2}{c}{32.9±2.9}&   \multicolumn{2}{c}{25.5±3.2}\\
3D-PointCapsNet~\cite{zhao20193d}&\multicolumn{2}{c}{42.3±5.5} & \multicolumn{2}{c|}{53.0±5.9} &\multicolumn{2}{c}{38.0±4.5}&   \multicolumn{2}{c}{27.2±4.7}\\
PointNet++~\cite{qi2017pointnet++}&\multicolumn{2}{c}{38.5±4.4} & \multicolumn{2}{c|}{42.4±4.5} &\multicolumn{2}{c}{23.1±2.2}&   \multicolumn{2}{c}{18.8±1.7}\\
PointCNN~\cite{li2018pointcnn}&\multicolumn{2}{c}{65.4±2.8} & \multicolumn{2}{c|}{68.6±2.2} &\multicolumn{2}{c}{46.6±1.5 }&   \multicolumn{2}{c}{50.0±2.3}\\
RSCNN~\cite{liu2019relation}&\multicolumn{2}{c}{65.4±8.9} & \multicolumn{2}{c|}{68.6±7.0} &\multicolumn{2}{c}{46.6±4.8}& \multicolumn{2}{c}{50.0±7.2}\\
\hline
&PNet&DGCNN&PNet&DGCNN&PNet&DGCNN&PNet&DGCNN\\\hline
Rand&52.0±3.8&31.6±2.8&57.8±4.9&40.8±4.6&46.6±4.3&19.9±2.1& 35.2±4.8&16.9±1.5\\
Jigsaw~\cite{sauder2019self}&66.5±2.5&34.3±1.3&69.2±2.4&42.2±3.5&56.9±2.5&26.0±2.4&66.5±1.4&29.9±2.6\\
cTree~\cite{sharma2020self}&63.2±3.4&60.0±2.8&68.9±3.0&65.7±2.6&49.2±1.9&48.5±1.8&50.1±1.6&53.0±1.3\\
PoinDisc~\cite{liu2021point}&67.8±2.3&81.5±1.8&71.7±2.9&83.3±2.3&53.4±2.7&68.1±3.0&56.3±2.2&70.5±2.9\\
PointContrast~\cite{xie2020pointcontrast}&69.2±2.9&83.6±2.4&72.5±2.5&87.4±2.8&51.7±3.1&71.5±2.5&57.9±2.4&74.5±3.2\\
OcCo~\cite{wang2021unsupervised}&89.7±1.9&90.6±2.8&92.4±1.6&92.5±1.9&83.9±1.8 &82.9±1.3&\textbf{89.7±1.5}&86.5±2.2\\
\textbf{CLR-GAM (ours)}&\textbf{91.8$\pm$2.6}&\textbf{93.7$\pm$1.2}&\textbf{94.8$\pm$2.4}&\textbf{96.0$\pm$2.6}&\textbf{84.6$\pm$2.2}&\textbf{87.9$\pm$2.7}&89.1$\pm$2.0&\textbf{91.1$\pm$1.9}\\\hline
CrossPoint~\cite{afham2022crosspoint}&90.9±4.8&92.5±3.0&   93.5±4.4&94.9±2.1&   84.6±4.7&83.6±5.3&   \underline{90.2±2.2}&87.9±4.2\\
\end{tabular}
}
\caption{Few shot object classification on ModelNet-40. A linear SVM is fit on the training set of ModelNet-40 using the pretrained model learned from ShapeNet. Compared with existing methods, the proposed CLR-GAM achieves state-of-the-art performance under different few-shot settings.
The results are the overall accuracy.}
\label{tbl:fsl_mn10}
\vspace{-0.2cm}
\end{table*}

\begin{table*}[t!]
\centering
\resizebox{0.8\textwidth}{!}{
\begin{tabular}{ l | c c c c | c c c c}
&\multicolumn{4}{c|}{5-way}&\multicolumn{4}{c}{10-way}\\\hline
Method &\multicolumn{2}{c}{10-shot} &\multicolumn{2}{c|}{20-shot} &\multicolumn{2}{c}{ 10-shot} &\multicolumn{2}{c}{20-shot}\\\hline\hline
&PNet&DGCNN&PNet&DGCNN&PNet&DGCNN&PNet&DGCNN\\\hline
Rand&57.6±2.5 &62.0±5.6& 61.4±2.4 &67.8±5.1& 41.3±1.3&37.8±4.3 & 43.8±1.9& 41.8±2.4\\
Jigsaw~\cite{sauder2019self}&58.6±1.9&65.2±3.8& 67.6±2.1&72.2±2.7& 53.6±1.7&45.6±3.1& 48.1±1.9&48.2±2.8\\
cTree~\cite{sharma2020self}&59.6±2.3&68.4±3.4& 61.4±1.4&71.6±2.9& 53.0±1.9&42.4±2.7& 50.9±2.1&43.0±3.0\\
PoinDisc~\cite{liu2021point}&57.4±1.8&58.2±2.7&58.1±2.7&60.1±2.1&38.1±2.5&41.5±2.9&39.5±2.2&40.1±3.1\\
PointContrast~\cite{xie2020pointcontrast}&59.2±1.7&60.6±3.2&62.7±2.9&65.5±2.6&42.3±1.7&46.3±3.2&45.9±1.7&52.5±2.9\\
OcCo~\cite{wang2021unsupervised}&70.4±3.3& 72.4±1.4&  72.2±3.0&77.2±1.4& 54.8±1.3& 57.0±1.3& 61.8±1.2&61.6±1.2\\
\textbf{CLR-GAM (ours)}&\textbf{71.8$\pm$2.8}&\textbf{80.6$\pm$1.9}&\textbf{78.4$\pm$3.2}&\textbf{86.3$\pm$2.3}&\textbf{63.8$\pm$2.6}&\textbf{67.2$\pm$1.5}&\textbf{69.4$\pm$2.8}&\textbf{76.4$\pm$2.7}\\
\hline
CrossPoint~\cite{afham2022crosspoint}&68.2±1.8&74.8±1.5&   73.3±2.9&79.0±1.2&   58.7±1.8&62.9±1.7&   64.6±1.2&73.9±2.2\\
\end{tabular}
}
%
\caption{Few shot object classification on ScanObjectNN. A linear SVM is fit on the training set of ModelNet-40 using the pretrained model learned from ShapeNet. Compared with existing methods, the proposed CLR-GAM outperforms state-of-the-art method~\cite{wang2021unsupervised} by a large margin. Reported results are the overall accuracy.}
\label{tbl:scanobject_fsl}
\vspace{-0.4cm}
\end{table*}

\noindent\textbf{b) Few Shot Object Classification:}
\label{subsec: few-shot}
Few Shot Learning (FSL) is a learning paradigm that aims to train a model that generalizes with limited data.
%
In this experiment, we conduct experiments on N-way K-shot learning, which means that a model is trained on N classes and K samples in each class.
The test/query set for each N class consists of 20 unseen samples for all these experiments. 
We use ModelNet-40 and ScanObjectNN 
for these experiments. 
The same pretrained model is used for both classification and FSL tasks with respective backbones. 
Similar to the classification task, we fit a linear SVM classifier for testing the FSL task. A similar protocol is used in earlier works~\cite{afham2022crosspoint, sharma2020self}. 
We report the results in Tables~\ref{tbl:fsl_mn10},~\ref{tbl:scanobject_fsl}. 
As there is no standard benchmark test set, we follow the setting used in~\cite{afham2022crosspoint, sharma2020self, wang2021unsupervised}. Specifically, we report mean and standard deviation over 10 runs. 

As shown in Table~\ref{tbl:fsl_mn10}, we observe that the CLR-GAM with DGCNN achieves SOTA compared to all other approaches in the challenging 5-way setting. 
In the 10-way setting, CLR-GAM performs on par with CrossPoint (multimodal pretrained) and Occo (single modal pretrained). The results show the same trend as in~\ref{tbl:classification_mn40_results}.
The few-shot object classification results on ScanObjectNN is reported in Table~\ref{tbl:scanobject_fsl}. %
CLR-GAM with DGCNN and PointNet performs SOTA compared to both point cloud and multimodal pretrained approaches. Specifically, on ScanNet we show a large margin improvement (more than 11\%) using  DGCNN on all sets, and more than 8\% improvement with PNET (5 way-20 shot, 10 way-10 shot, 10 way-20 shot). There is a 24\% improvement with both DGCNN and PNET backbones in 10 way-20shot.   
The results further testify that CLR-GAM learns discriminative 3D point cloud representations, and the representations can generalize to challenging real-world settings. 

\begin{figure*}[t!]
    \centering
    \includegraphics[width=0.8\textwidth]{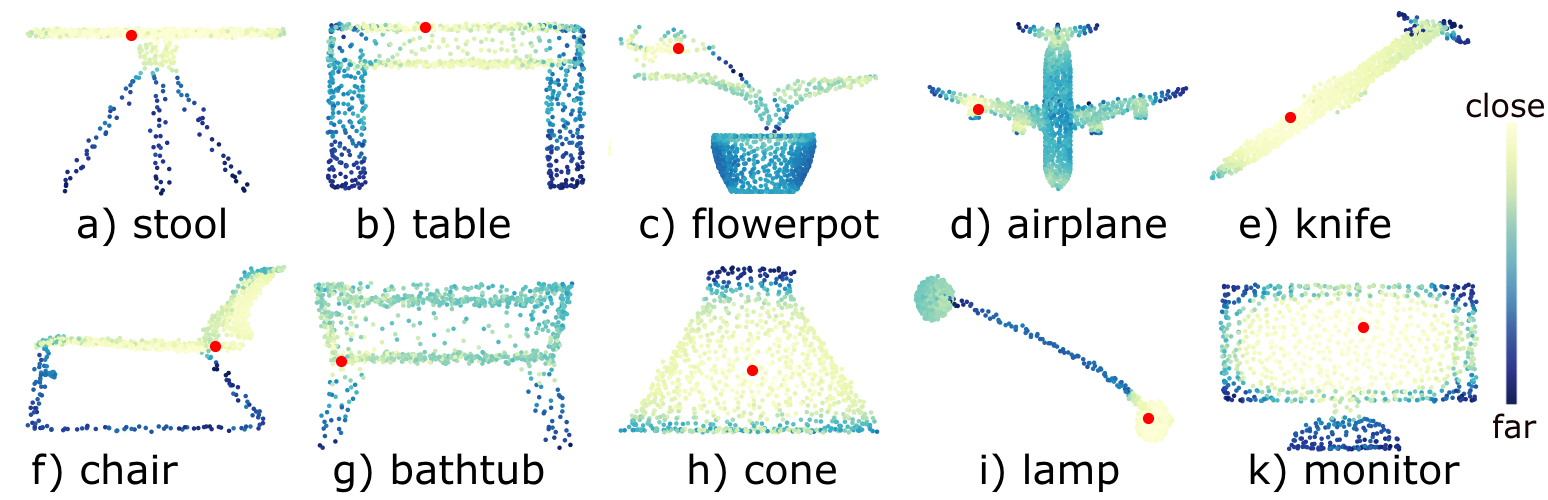}
    \caption{Feature visualization of unseen objects selected from the test sets of ShapeNet and ModelNet-40. For more qualitative results please check the supplementary material.
    }
    \label{fig:feat_viz}
    \vspace{-0.5cm}
\end{figure*}
\noindent\textbf{c) 3D Object Part Segmentation:}
\begin{table}[h!]
\centering
\resizebox{0.85\columnwidth}{!}{
\begin{tabular}{ c|c|c } 
Category & Method & Mean IOU \\\hline
Supervised & PointNet~\cite{qi2017pointnet} & 83.7\\
 & PointNet++~\cite{qi2017pointnet++} & 85.1\\
 & DGCNN~\cite{wang2019dynamic} & 85.1\\
\hline
Self-Supervised & Self-Contrast~\cite{du2021self} & 82.3\\
 & PointContrast ~\cite{xie2020pointcontrast} & 85.1\\
 & PointDisc~\cite{liu2021point} & 85.3\\
 & Jigsaw~\cite{sauder2019self} & 85.3\\
 & OcCo~\cite{wang2021unsupervised} & 85.0\\
 & \textbf{CLR-GAM (ours)} & \textbf{85.5} \\\hline
+ Image Modality & CrossPoint~\cite{afham2022crosspoint} & 85.5\\
\end{tabular}
}
\caption{We report the mean IOU results for 3D object part segmentation on the ShapeNet-part dataset. Supervised methods are trained with randomly initialized weights, whereas self-supervised methods are initialized with pretrained weights learned from  ShapeNet.}
\label{tbl:part_seg}
\vspace{-0.3cm}
\end{table}
ShapeNet-part dataset~\cite{yi2016scalable}, which contains 50 different parts from 16 distinct object categories with a total of 16,881 3D objects, is used for 3D part object segmentation. 
We use the same pretrained model for both classification and FSL tasks with respective backbones. 
To be consistent with the evaluation for part segmentation, we finetune the pretrained model using 2048 points sampled from point clouds. 
We observe that the performance of CLR-GAM is better than the other point cloud contrastive learning-based approaches and on par with CrossPoint (multimodal pretrained). 
The reported results in Table~\ref{tbl:part_seg} are the average of intersection over union (IOU) computed for each part.
\begin{table}[h!]
\centering
\resizebox{\columnwidth}{!}{
\begin{tabular}{ c|c|c|c|c|c|c|c } 
\multicolumn{5}{c|}{augmentations}&\multicolumn{2}{c|}{novel modules}&dataset\\\hline
jitter&translation & rotation & scaling & crops & GFM& GA& Modelnet-40\\
\hline
\hline
\checkmark&\checkmark&\checkmark &\checkmark&&&&84.8\\ 
\checkmark&\checkmark&\checkmark &\checkmark&\checkmark&&&89.7\\ 
\checkmark&\checkmark&\checkmark &\checkmark&\checkmark&\checkmark &&90.7\\
\checkmark&\checkmark&\checkmark &\checkmark&\checkmark&&\checkmark& 90.4\\
\checkmark&\checkmark&\checkmark &\checkmark&\checkmark&\checkmark&\checkmark&\textbf{91.3}\\ 
\end{tabular}
}
\caption{Ablation Study of CLR-GAM: Trained on ShapeNet using the self-supervised method and evaluated ModelNet-40 using Linear-SVM. Reported results are overall accuracy }
\label{tbl:ablation_augs}
\vspace{-0.5cm}
\end{table}
\subsection{Qualitative Results}
We visualize feature representations (learned from the proposed CLR-GAM) of each point/node in an unseen object's point cloud selected from test sets of ShapeNet and ModelNet-40 in Figure~\ref{fig:feat_viz}. 
We compute the cosine distance between the feature of a randomly selected point (colored in red) to other points' features in the same point cloud. 
The color scale is Yellow-Green-Blue. The closest feature in the feature space is yellow, and the farthest is blue.
Our approach learns similar representations for the whole planar region for simple planar structures such as stool (a) and table (b).  
Moreover, in the case of a chair (f), a complicated planar structure, the proposed model can learn similar features for the back part of a seat.
For monitor (k), the plane is assigned with closer/similar features, whereas the features at the corners (structural irregularities) are dissimilar to the center. 
A similar observation can be found in the case of a knife (e), i.e., the handle and sharp edge have different features.  
For a curved object like a bathtub (g), the whole tub has similar features except for the legs. Similarly, for the cone (h), the whole curved region has similar features except for the edges. In the case of lamp (i), the curved stand has similar features, separating the stem. 
For irregular-shaped objects, e.g., flowerpot (c), all leaves have similar features, and different features are learned for pot and stem. 
For airplane (d), all turbines have similar features since it is relatively small and curved, and the other sharply curved front and back regions of the airplane have similar features.      
\begin{figure*}[t!]
    \centering
    \includegraphics[width=0.9\textwidth]{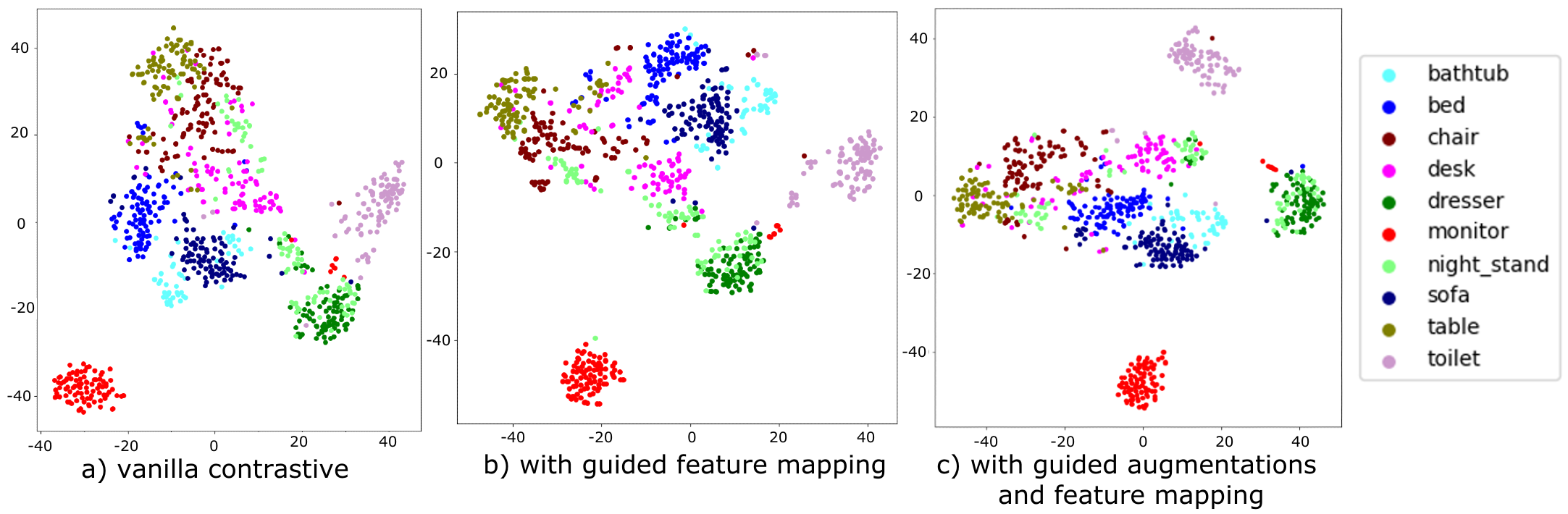}
    \vspace{-0.4cm}
    \caption{t-SNE plots: visualization of features from three different approaches, generated from unseen samples of ModelNet-10 test dataset.
    }
    \label{fig:tsne_plots}
    \vspace{-0.4cm}
\end{figure*}
\subsection{Ablation Study}
\noindent\textbf{a) Augmentations and Novel Modules: }We conduct an ablation study on the ModelNet-40 dataset to understand the contribution of GFM, GA, and augmentation. The results are shown in Table~\ref{tbl:ablation_augs}. 
Contrastive learning without cropping achieves around 84.8\% in overall accuracy. 
With cropping, a large improvement of 4.9\% is observed. 
The result is similar to the performance of CrossPoint~\cite{afham2022crosspoint} without multimodal training (i.e., only Intra Modal Instance Discrimination, IMID). 
We treat the model as the vanilla baseline, i.e., the second row in Table~\ref{tbl:ablation_augs}. 
With GFM, we observe a performance improvement of 1.1\% compared to the vanilla baseline. A 0.77\% improvement is observed when GA is added. 
%
%
When both novel modules are introduced, we observe a 1.78\% improvement compared to the vanilla baseline.
The ablative studies demonstrate the effectiveness of the proposed GA and GFM.

\noindent\textbf{b) Feature Visualization:} We depict all features generated from our CLR-GAM approach on unseen samples of the ModelNet-10 test dataset using the DGCNN backbone in Figure~\ref{fig:tsne_plots}. 
To generate t-SNE plots, we use a perplexity of 30. 
In the vanilla contrastive learning approach, except monitor class, all the other classes have a wider spread making the classes closer. 
With the proposed GFM, we observe the improvement in nightstand toilet classes, but with a similar overlap of bed bathtub classes as vanilla. 
With added GA, our proposed approach CLR-GAM, we observe further improvement in toilet class separation from the nightstand and more concentrated class clusters. 
In all cases, the dresser and nightstand were more confused because of the shape similarity. 

\noindent\textbf{c) Generalization to driving scene:} To understand the generalization of the proposed unsupervised approach to the real-world applications or datasets, we visualize the features of a driving scenario point-cloud data from the KITTI dataset~\cite{geiger2013vision}, which is shown in Figure~\ref{fig:feat_viz_kitti}. The full scene with 80 meters in all directions of the ego-vehicle (160m x 160m scene) is shown in (a) as a top-down image. In subfigure-a, the gray color is used for the ground, and the red color is used for the non-ground or obstacles. The separation is done using -1.5 meters in height axis of the point cloud data from the Velodyne sensor. The blue box is the region of interest (ROI), which is zoomed in subfigure-b, which is a 20m x 20m region. This is subsampled to around 4000 points using voxel-based sampling with a 0.3-meter voxel length in all three axes. 1024 points are randomly selected and passed to the feature encoder. The features are visualized in subfigure-c. The color scale is the same as in Figure~\ref{fig:feat_viz}, Yellow-Green-Blue. The closest feature in the feature space is yellow, and the farthest one is blue with respect to a randomly selected point (red circle). The two vehicles have features different from the ground, highlighted in pink boxes.  
\section{Discussion}
\begin{figure}[h!]
\centering
\includegraphics[width=\columnwidth]{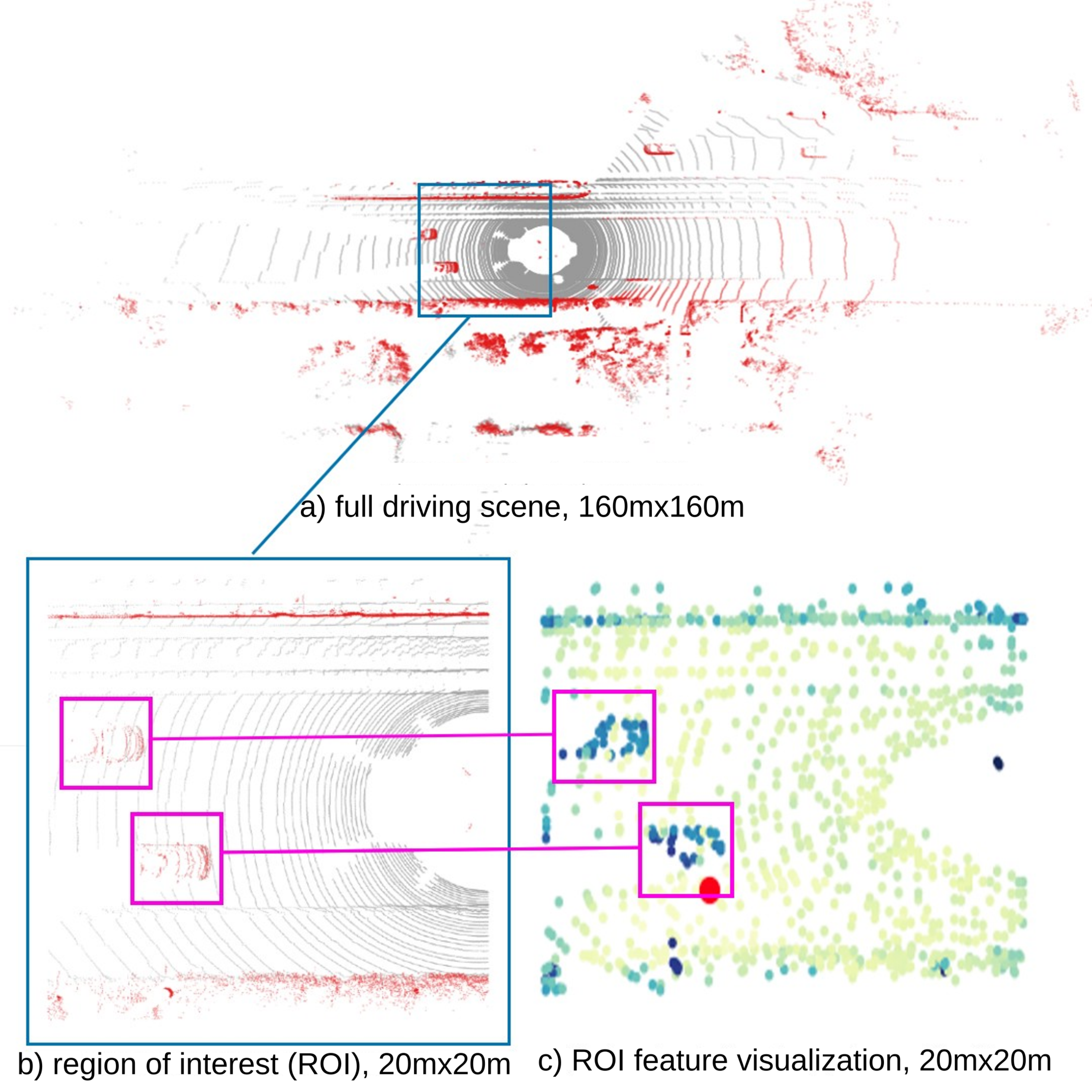}%
\caption{Feature visualization of unseen \textbf{driving scene} selected from the KITTI dataset.}
\label{fig:feat_viz_kitti}
\vspace{-0.5cm}
\end{figure}

Existing local feature contrastive techniques using inter point cloud features (PointContrast)~\cite{xie2020pointcontrast} or intra point cloud features (PointDisc)~\cite{liu2021point} objective suffers from LCA. It is also observed from our qualitative results shown in Figure~\ref{fig:feat_viz} that similar parts/shapes or symmetries in an object can have similar features. Compared to these local contrast approaches, our novel approach performs better in downstream tasks using linear SVM on the learned feature representation. Our proposed approach avoids LCA by avoiding the local feature contrast objective. But global contrast introduces GCA as mentioned in Section~\ref{sec:intro}. With our novel GFM and global contrast, we address GCA and also perform better than other global contrast techniques DepthContrast, pretext-based approaches, and multimodal trained CrossPoint from our quantitative evaluation. Our proposed self-supervised approach not only generalizes to different object datasets but also to driving scenes, as shown in Figure~\ref{fig:feat_viz_kitti}. Please check the supplementary for further discussions.
\section{Conclusion}
In this paper, we present a contrastive learning framework (CLR-GAM) with guided augmentation (GA) to search augmentation parameters efficiently and guided feature mapping (GFM) to associate structural features precisely. The former is realized by adapting the inverse Dirac delta function with a memory bank, and the latter is fulfilled by the global contrasting of associated structural features between two augmented point clouds. Both these processes help boost the contrastive learning of point cloud data. We benchmark on three different downstream tasks and show that our method performs state-of-the-art compared to other methods trained on single modality point cloud data. It also performs similar to or better than a recent multimodal trained approach CrossPoint.
{\small
\bibliographystyle{ieee_fullname}
\bibliography{egbib}
}
\newpage
\phantom{a}
\begin{figure*}[t]
\begin{center}
 \includegraphics[width=\textwidth]{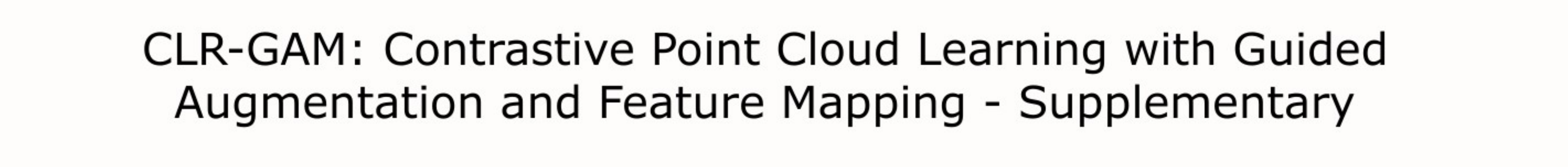}
\label{fig:caption_supp}
    \vspace{-0.8cm}
\end{center}
\end{figure*}
\newpage
\phantom{a}
\section{Pretraining}
We pretrain on the Shapenet dataset~\cite{chang2015shapenet} using the proposed contrastive self-supervised approach, similar to other reported benchmarks~\cite{afham2022crosspoint, sauder2019self, huang2021spatio}. 
The dataset has 55 different classes with a total of 57,386 CAD models. 
We sample 2048 points before performing augmentation and 1024 after applying all augmentations as mentioned in section 3.2.b. 
The pretrained model is benchmarked on three downstream tasks. 

\section{Implementation Details}
For all of the experiments, we use cyclic learning rates~\cite{smith2017cyclical} for 3 cycles with each cycle for 100 epochs and a cosine annealing based learning scheduler. 
We employ Adam optimizer with a learning rate of $10^{-3}$ and a weight decay of $10^{-4}$. 
PNet~\cite{qi2017pointnet} and DGCNN~\cite{wang2019dynamic} are utilized for point cloud encoding.
%
We apply augmentation ranges for translation, rotation, and scaling as mentioned in section 3.2.a. For jittering, we apply Gaussian noise of 1cm standard deviation. 
For cropping, we randomly select a point and crop 30\% of the points that are  closer to the selected point.
%
Three downstream tasks are benchmarked in this paper, i.e., classification, few-shot learning, and object part semantic segmentation.
For training, we use two NVIDIA RTX 6000 GPUs with a batch size $B$ of 32. For Equation 2 in the main manuscript, we use $\alpha_R=1,\alpha_T=1,\alpha_S=1$. 

\section{Guided Augmentation sampling vs Random sampling}
\begin{figure}[h!]
    \centering
    \includegraphics[width=0.8\columnwidth]{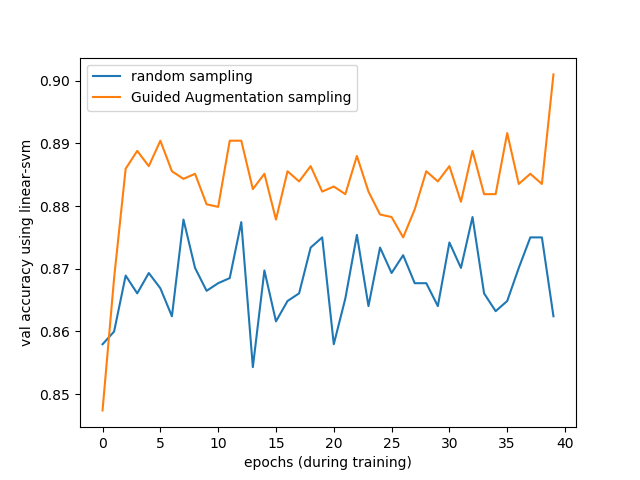}
    \caption{Validation Accuracy on ModelNet-40 using LinearSVM, during self-supervised training with random and guided augmentation sampling.
    }
    \label{fig:convergence}
\end{figure}

We visualize the convergence in performance over 40 epochs for two different sampling techniques in Figure~\ref{fig:convergence}. During self-supervised training on the ShapeNet dataset, the performance (accuracy) is evaluated on the validation set of ModelNet-40 after every epoch.

The standard deviation for Guided Augmentation from Table 6 under multiple runs (5) is +/-0.12, compared to the random selection process +/-0.37

\section{Limitations} 
The proposed GA module uses a very effective memory mechanism, but it might not be memory efficient with many augmentation samples. 
It takes 3 minutes and 30 seconds for 35000 augmentations (around the sample size of shapenet dataset), without any advanced libraries (only using the NumPy library with naive implementation) and the storage memory footprint is 2.52 MB (with 8 bytes per element in the array). Please note that we train linear-SVM on the features on tasks (classification/few-sot learning) for both datasets (ModelNet-40/ScanNet), because of this the memory limitation only applies to the pretrained dataset. 
\section{Discussions}
\subsection{Memory size on different datasets}
We trained only one dataset for self-supervised learning (ShapeNet dataset) even though there are different tasks/datasets that are tested using Linear-SVM. So, in our experiments, it does not change with tasks/datasets that are tested on. But without memory, there is a performance degradation of 0.8\%, as seen in Table 7. We chose memory based on the size of the data set that is pre-trained on.
\subsection{Why Guided Feature Mapping when there is a pooling operation?}
The pooling operation is performed on the encoded features and before latent feature projection. But because of cropping the same point cloud can resemble being coming from two different classes, as mentioned in the Introduction section. So we hypothesize that only pooling features that have similar structural similarities will result in an effective contrastive learning, which is also observed in our empirical results. To study the effectiveness of the Latent features, in the main manuscript we also show t-SNE plots in Figure 4.
\subsection{Extension to other sensor modalities}
This is an interesting future direction that can be explored. Based on our understanding, our approach can be applied to such works, as crosspoint. To ensure efficient cross-modal embedding, we also need to search for the right approaches for images. That is not the focus of this paper, so we leave it to future work.

\section{Qualitative Results (ModelNet40)}
We visualize feature representations (learned from the proposed CLR-GAM) of each point/node in an unseen object’s point cloud selected from test sets of ModelNet-40 in Figure 3. The color scale is same as Figure 3 in main manuscript, Yellow-Green-Blue. The closest feature in the feature space is yellow, and the farthest is blue with respect to a selected point (colored in red). Some qualitative results and discussions of the airplane, bathtub, bed, guitar, person, vase and lamp are shown below.  
\subsection{airplane}
In the Figure~\ref{fig:feat_viz_airplane}(a-d) we visualize four different airplanes pointcloud features. In (a,b,d) the selected points (red dot) for the three different planes are on the wings. Except the sharper wings ends or tail ends or engines or mouth of the airplane, the whole body of the plane has similar features. Similarly, in (c) when selected sharper wing end (red dot), tail wings are more closer in the feature space, along with engines and mouth of the airplane.
\begin{figure}[h]
    \centering
    \includegraphics[width=\columnwidth]{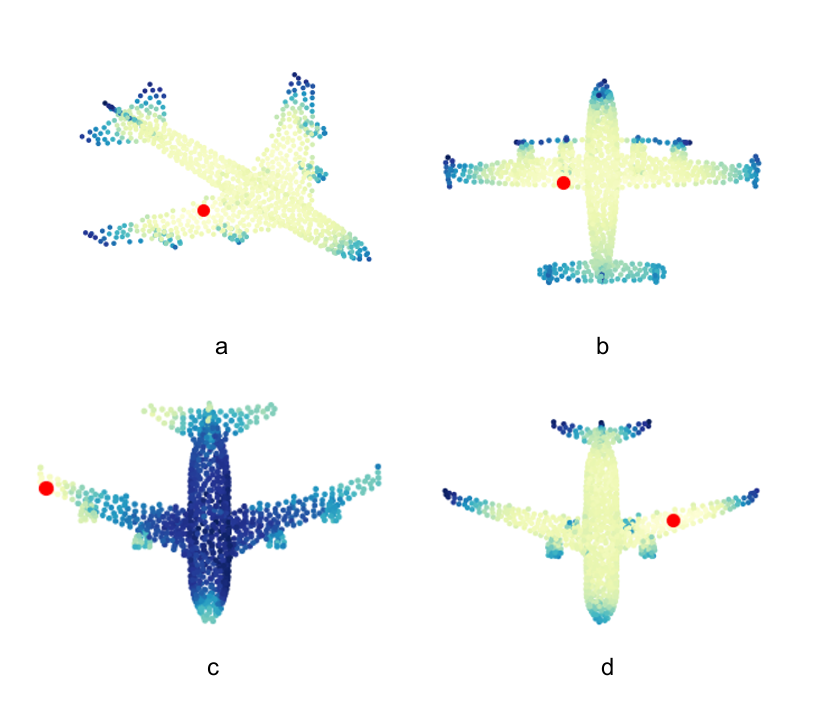}
    \caption{Feature visualization of unseen \textbf{(airplane)} objects selected from the test sets of ModelNet-40.
    }
    \label{fig:feat_viz_airplane}
\end{figure}
\subsection{bathtub}
\begin{figure}[h]
    \centering
    \includegraphics[width=\columnwidth]{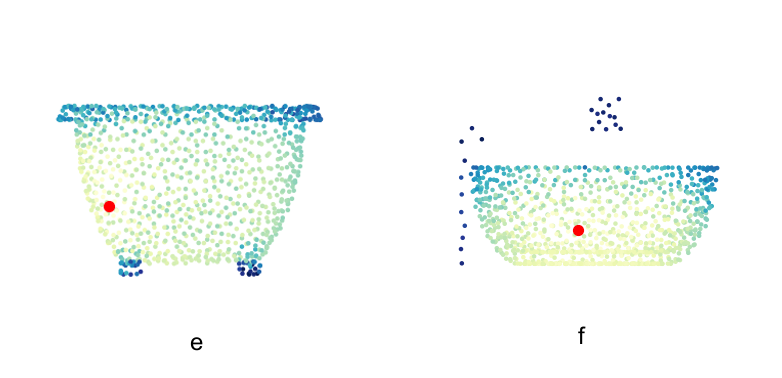}
    \caption{Feature visualization of unseen \textbf{(bathtub)} objects selected from the test sets of ModelNet-40.
    }
    \label{fig:feat_viz_bathtub}
\end{figure}

In the Figure~\ref{fig:feat_viz_bathtub}(e,f) we visualize two different bathtub pointcloud features. In (e,f) we selected points shown in red dot are on the tub. In (e) the whole symmetrical tub shape has similar features excluding the legs and top edge handle. Similarly in (f) the tap/handle,  separate object and sharp corners has different features from the rest of the bath tub.  
\subsection{bed}
\begin{figure}[h]
    \centering
    \includegraphics[width=\columnwidth]{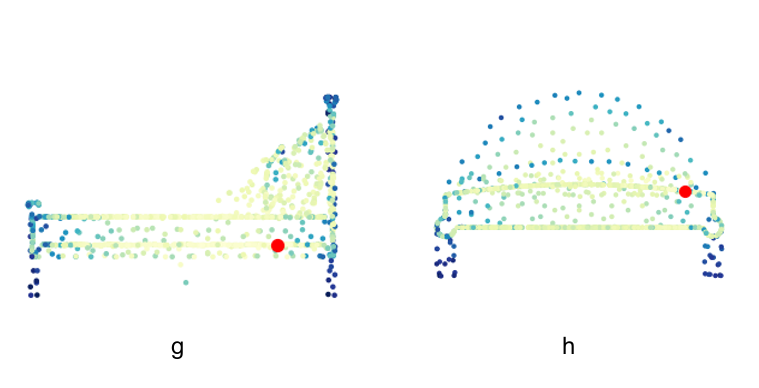}
    \caption{Feature visualization of unseen \textbf{(bed)} objects selected from the test sets of ModelNet-40.
    }
    \label{fig:feat_viz_bed}
\end{figure}

In the Figure~\ref{fig:feat_viz_bed}(g,h) we visualize two different bed pointcloud features. In (g) the selected point (red dot) is on box spring, the whole part has similar features excluding legs and head board . In (h) the selected point is close to foot board, since there is no separate foot board in this pointcloud the whole box spring has similar features excluding legs and head board.
\subsection{guitar}
\begin{figure}[h]
    \centering
    \includegraphics[width=\columnwidth]{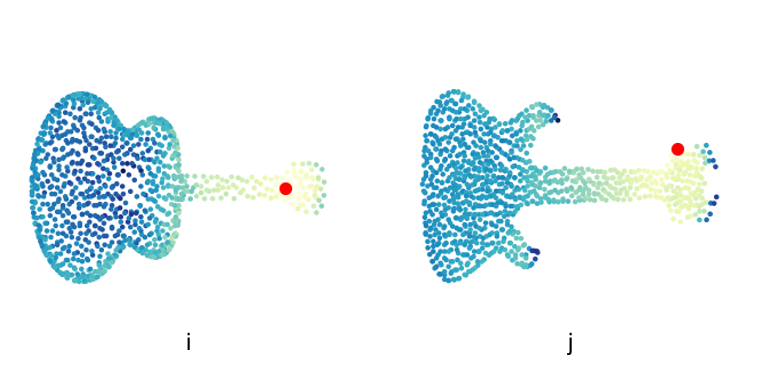}
    \caption{Feature visualization of unseen \textbf{(guitar)} objects selected from the test sets of ModelNet-40.
    }
    \label{fig:feat_viz_guitar}
\end{figure}

In the Figure~\ref{fig:feat_viz_guitar}(i,j) we visualize two different guitar pointcloud features. In (i) the selected point (red dot) is on the nut, the whole finger board and head stock has same features excluding the body (since the head stock doesn't have any varied design as shown in (j)). In (j) the selected point is on head stock, only head stock and nut has similar features, finger board and body have different features.
\subsection{person}
\begin{figure}[h]
    \centering
    \includegraphics[width=\columnwidth]{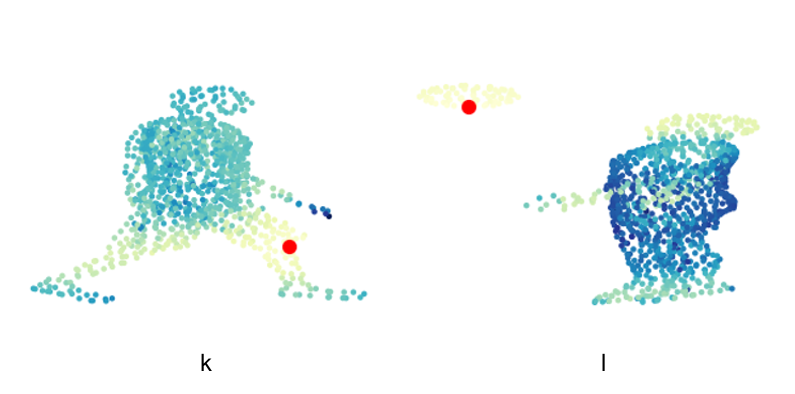}
    \caption{Feature visualization of unseen \textbf{(person)} objects selected from the test sets of ModelNet-40.
    }
    \label{fig:feat_viz_person}
\end{figure}
In the Figure~\ref{fig:feat_viz_person}(k,l) we visualize two different person pointcloud features. In (k) the selected point (red dot) is on the leg, both the legs have same features excluding the feet and the upper body. Similary in (l) the selected point is on the ball and the person is catching the ball in this pointcloud. The person's head and the ball have same features because they are round in shape.   

\subsection{vase}
\begin{figure}[h]
    \centering
    \includegraphics[width=\columnwidth]{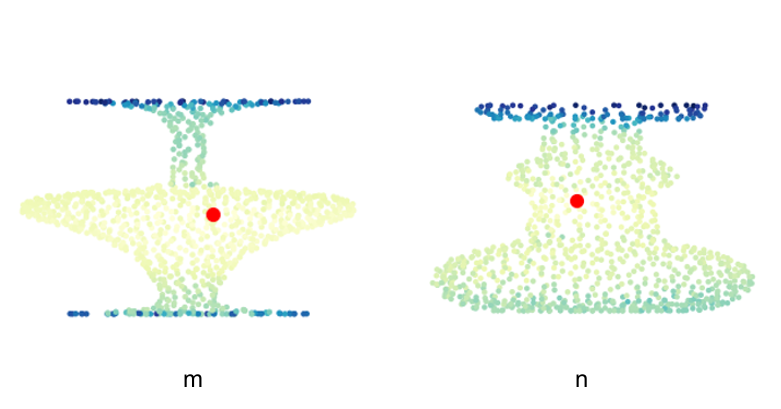}
    \caption{Feature visualization of unseen \textbf{(vase)} objects selected from the test sets of ModelNet-40.
    }
    \label{fig:feat_viz_vase}
\end{figure}

In the Figure~\ref{fig:feat_viz_vase}(m,n) we visualize two different vase pointcloud features. In (m) the selected point (red dot) is on the body of the vase, the whole body has similar features excluding the lip, foot and neck. In (n) the selected point is also on the body. Even though the body shape is complicated the whole body has similar features, excluding the lip.    
\subsection{lamp}
\begin{figure}[h]
    \centering
    \includegraphics[width=\columnwidth]{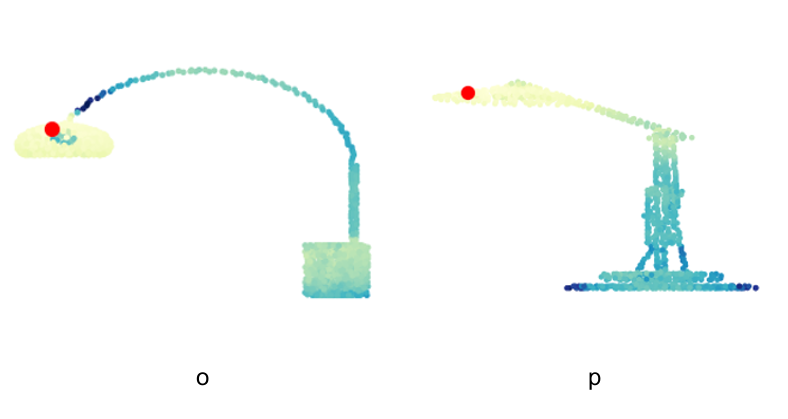}
    \caption{Feature visualization of unseen \textbf{(lamp)} objects selected from the test sets of ModelNet-40.
    }
    \label{fig:feat_viz_lamp}
\end{figure}
In the Figure~\ref{fig:feat_viz_lamp}(o,p) we visualize two different lamp pointcloud features. In both cases the selected point (red dot) is on the shade. In (o) the complete shade has same features, even tough the bulb and tube are closer they have different features. In case of (p) the shade and bridge arm have same features, excluding the base and tube.
\vspace{0.5cm}
\section{Notations}
\centerline{\bf Augmentations}
\bgroup
\def\arraystretch{1.5}
\begin{tabular}{p{0.2in}p{3.25in}}
$\displaystyle \mathbf{a}$ & set of all augmentations\\
$\displaystyle \mathbf{a}^S$ & scaling \\
$\displaystyle \mathbf{a}^T$ & translation \\
$\displaystyle \mathbf{a}^R$ & rotation \\
$\displaystyle \mathbf{a}^J$ & jitter \\
$\displaystyle \mathbf{a}^C$ & crop \\
$\displaystyle \mathbf{\hat{a}}$ & randomly sampled augmentation \\
$\displaystyle \mathbf{a^*}$ & novel augmentation \\
$\displaystyle \mathbf{a^{-1}}$ & inverse augmentation \\
\end{tabular}
\egroup
\vspace{1.2cm}

\newpage
\centerline{\bf PointCloud, Features, Memory}
\bgroup
\def\arraystretch{1.5}
\begin{tabular}{p{0.2in}p{3.25in}}
$\displaystyle P$ & pointcloud\\
$\displaystyle \mathbf{x}$ & points in pointcloud \\
$\displaystyle n$ & number of points in the pointcloud \\
$\displaystyle \mathbb{R}$ & real numbers \\
$\displaystyle z$ & projected latent feature \\
$\displaystyle N$ & number of randomly sampled augmentations \\
$\displaystyle M$ & memory bank \\
$\displaystyle m$ & index of the memory slot in memory bank \\
$\displaystyle K$ & dirac delta kernal function \\
$\displaystyle d$ & total distance measure \\
$\displaystyle d_R$ & angular distance measure \\
$\displaystyle c,\epsilon$ & small values for numerical stability \\
$\displaystyle S$ & structural index mapping\\
$\displaystyle \otimes$ & augmentation operator\\
$\displaystyle \odot$ & indexing operator\\
$\displaystyle B$ & size of mini-batch\\
$\displaystyle b$ & index of the feature in mini-batch\\
\end{tabular}
\vspace{0.2cm}
\newpage
\centerline{\bf Indexing}
\bgroup
\def\arraystretch{1.5}
\begin{tabular}{p{0.2in}p{3.25in}}
$\displaystyle P_i$ & sample $i$ of pointcloud from the dataset \\
$\displaystyle F_i$ & feature set corresponding to the sample $i$ of pointcloud \\
$\displaystyle \mathbf{x}_{j}$ & $j$th point in the pointcloud \\
$\displaystyle F(j)$ & feature corresponding to the $j$th point in the pointcloud \\
$\displaystyle \mathbf{a}_{k}$ & $k$th augmentation  \\
$\displaystyle P^k$ & pointcloud augmented with augmentation with index $k$ \\
$\displaystyle F^k$ & features of pointcloud with augmentation with index $k$ \\
$\displaystyle S_{12}$ & structural index mapping from pointcloud 1 to 2\\
\end{tabular}
\egroup
\end{document}